\documentclass{article}

\usepackage{arxiv}

\usepackage[utf8]{inputenc} 
\usepackage[T1]{fontenc}    
\usepackage{hyperref}       
\usepackage{url}            
\usepackage{booktabs}       
\usepackage{amsfonts}       
\usepackage{nicefrac}       
\usepackage{microtype}      
\usepackage{lipsum}
\usepackage{graphicx}
\usepackage{amsmath}
\graphicspath{ {./images/} }

\newcommand{\mC}{{\bf{C}}}
\newcommand{\mM}{{\bf{M}}}
\newcommand{\mSigma}{{\bf{\Sigma}}}
\newcommand{\mU}{{\bf{U}}}
\newcommand{\mV}{{\bf{V}}}
\newcommand{\R}{\mathbb{R}}
\newcommand{\E}{\mathbb{E}}

\newcommand\blfootnote[1]{%
  \begingroup
  \renewcommand\thefootnote{}\footnote{#1}%
  \addtocounter{footnote}{-1}%
  \endgroup
}

\title{Out-of-Distribution Example Detection in Deep Neural Networks using Distance to Modelled Embedding}

\author{
 Rickard Sjögren \\
  Sartorius Corporate Research \\
  Umeå, Sweden \\
  \texttt{rickard.sjoegren@sartorius.com} \\
   \And
 Johan Trygg \\
  Sartorius Corporate Research \\
  Umeå, Sweden \\
  \texttt{johan.trygg@sartorius.com} \\ \\
  Computational Life Science Cluster (CLiC) \\ 
  Department of Chemistry \\
  Umeå University \\
  Umeå, Sweden \\
  \texttt{johan.trygg@umu.se} \\
  
}

\begin{document}
\maketitle
\begin{abstract}
Adoption of deep learning in safety-critical systems raise the need for understanding what deep neural networks do not understand after models have been deployed. The behaviour of deep neural networks is undefined for so called out-of-distribution examples. That is, examples from another distribution than the training set. Several methodologies to detect out-of-distribution examples during prediction-time have been proposed, but these methodologies constrain either neural network architecture, how the neural network is trained, suffer from performance overhead, or assume that the nature of out-of-distribution examples are known a priori. We present Distance to Modelled Embedding (DIME) that we use to detect out-of-distribution examples during prediction time. By approximating the training set embedding into feature space as a linear hyperplane, we derive a simple, unsupervised, highly performant and computationally efficient method. DIME allows us to add prediction-time detection of out-of-distribution examples to neural network models without altering architecture or training while imposing minimal constraints on when it is applicable. In our experiments, we demonstrate that by using DIME as an add-on after training, we efficiently detect out-of-distribution examples during prediction and match state-of-the-art methods while being more versatile and introducing negligible computational overhead.
\blfootnote{
    A Pytorch implementation of DIME is available at: \url{https://github.com/sartorius-research/dime.pytorch}
}
\end{abstract}



\section{Introduction}
Thanks to the powerful transformations learned by deep neural networks, deep learning is increasingly applied throughout society impacting many aspects of modern life. Due to their strong performance in many tasks, deep neural networks are increasingly used in safety critical systems, with vision systems for autonomous cars as a well-known example. But deep neural networks, as all data-driven models, have undefined behaviour when input data differs from training data. Adopting deep learning in increasingly complex and possibly safety-critical systems makes it crucial to know not only whether the model's predictions are accurate, but also whether the model should predict at all. If the model is able to detect so called out-of-distribution (OOD) observations during prediction directly, the system can fall back to safe behaviour minimizing negative consequences of faulty predictions. A tragic real-life example of possible consequences is the fatal collision of a pedestrian and an autonomous vehicle in March 2018 \cite{ntsb2018}. In the moments before the crash, the car's vision system erratically classified the pedestrian as different classes of object moving at different speeds in different frames, a possible symptom of OOD-examples. To avoid similar future accidents it is important to understand the limits of our models' learned representations in order to detect when observations are not recognized, so that autonomous decision making based on deep learning can be improved. 

\begin{figure}[t]
    \centering
    \includegraphics[width=\textwidth]{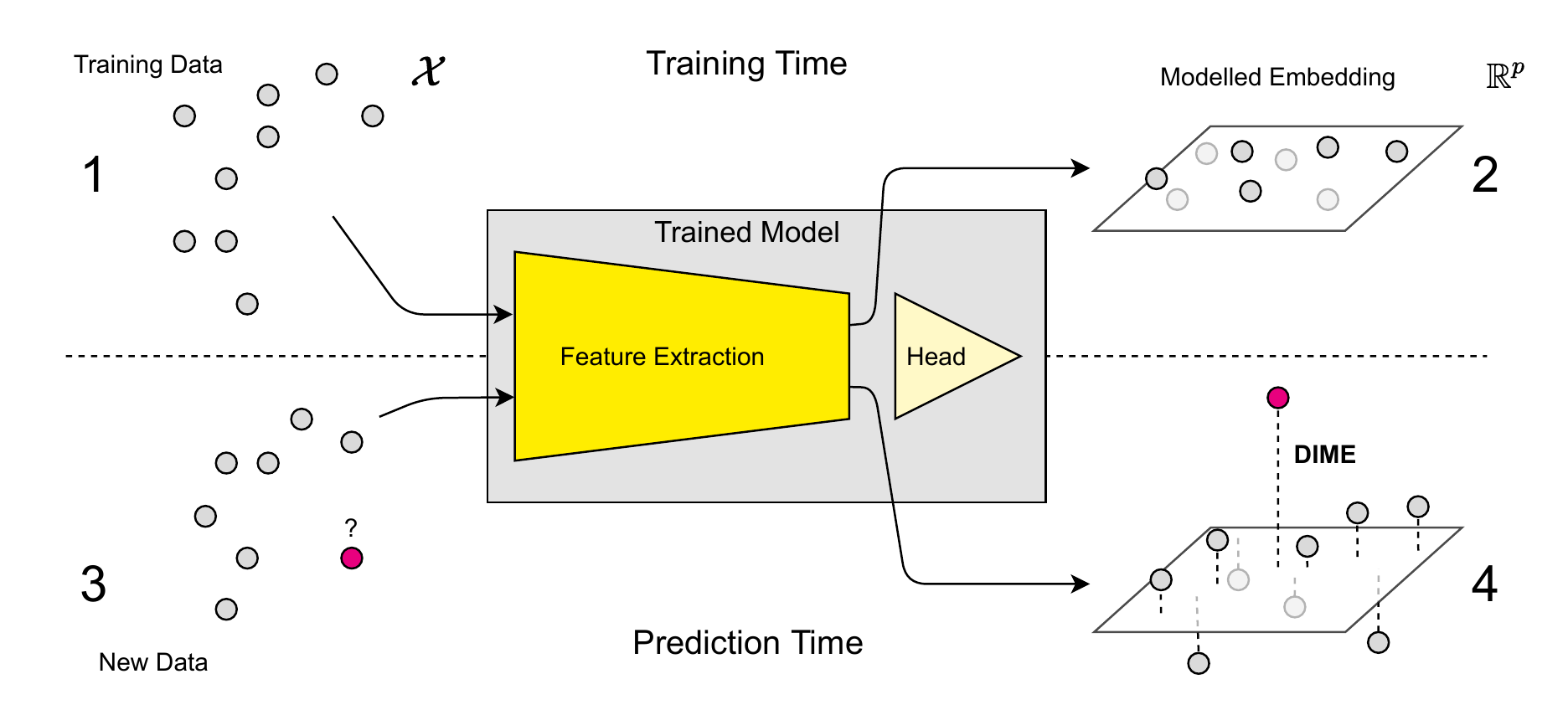}
    \caption{Overview of proposed workflow. Given a trained neural network, we transform training data drawn from some high-dimensional distribution in data space $\mathcal{X}$ into the model's intermediate feature vector space $\R^p$ (1). We linearly approximate the training set embedding as a hyperplane (2). When we then receive new observations it is difficult to assess if observations are out-of-distribution directly in data space, so we transform them into the same intermediate feature space (3). Finally, using the Distance-to-Modelled-Embedding (DIME) we can assess whether new observations fit into the expected embedding covariance structure (4). }
    \label{fig:workflow}
\end{figure}

Detecting OOD-examples in neural networks is a highly active field of research that is related to but different from anomaly detection where a separate model is fit to explicitly detect outliers from the training distribution. A practical solution to OOD-example detection is to filter input to a predictive model using an anomaly detection model, an approach that has been used for decades \cite{bishop1994novelty}. The downside of this approach is that it adds overhead and complexity by introducing another, possibly large, model. Another downside in the case of deep neural networks, is that there are no guarantees that the anomaly detection model captures the same variation that make the predictive model fail. Although the two approaches are complementary, it is preferable for a predictive model to be able to warn when it cannot handle input observations in a sensible manner.

A simple baseline for OOD-detection in neural networks was provided by Hendrycks and Gimpel \cite{hendrycks2016baseline} that observed that softmax-based classifiers often have low confidence for OOD-examples predictions. This is however far from generally true since neural networks are often poorly calibrated, meaning they may be very confident even for far OOD-examples. Several methods to improve calibration of neural networks are available, for instance relying on Platt-scaling \cite{guo2017calibration}, using temperature scaling and adding gradient noise to the input to increase confidence and indirectly increasing the confidence difference between inliers and OOD-examples \cite{liang2018enhancing}, or OOD-aware training using adversarially selected examples \cite{Li_2020_CVPR}. Albeit important, calibration is arguably not enough to provide a general solution by using the same scalar for both prediction and OOD-detection. For model outputs interpreted as probabilities it may suffice, but in the general case it is not clear how to distinguish between predicted amplitude and non-conformity of new observations.

A more general solution is to use a separate metric to detect OOD-examples. For instance, distance-based methods provide simple and effective means by measuring distance to training set observations either in data or feature space. For instance, class-conditional Mahalanobis distance in feature space of neural networks \cite{lee_unified_2018} achieved state-of-the-art performance in classification models. Another popular approach is to adapt neural networks to predict probability distributions rather than making point inferences, which can then be used to detect OOD-examples based on predictive uncertainty. To enable predictive uncertainty, methods have been formulated within a Bayesian framework for decades \cite{denker1991transforming,mackay1992practical} and methods published in recent years include:  sampling based on prediction time dropout \cite{gal_dropout_2016}, batch-normalization parameters \cite{teye2018bayesian}, model ensembles \cite{lakshminarayanan_simple_2017}, multiple prediction heads in a shared base network \cite{osband_deep_2016,ilg2018uncertainty,mandelbaum2017distance}, variational inference of weight distribution instead of regular point weights \cite{blundell2015weight} and Laplace approximation of distributions from existing weights \cite{ritter_scalable_2018}. But these Bayesian approaches constrain either how the network is constructed or how the network is trained. Many of them also rely on multiple forward-passes per prediction limiting their use in real-time systems or systems with limited computational resources.

Furthermore approaches include training linear "probe" classifiers to classify the target output given intermediate layers of a given base model that are fed to a meta-classifer trained to estimate whether or not the base model is correct \cite{chen2018confidence}. This method is however only formulated in terms of classification. Another approach explicitly modifies the training objective to differentiate between known outliers and training data to generalize to unknown examples \cite{hendrycks2018deep}, a procedure requiring knowledge of the nature of OOD-examples. Other work leverage generative methods, such as example likelihood based on an auto-regressive model \cite{ren2019likelihood} or even training a generative adversarial network \cite{goodfellow2014generative} to generate border-line outliers to allow calibration of a neural net classifier \cite{lee2017training}. Relying on generative models add significant complexity and requires modified model training procedures in order to be used.

To provide a method that reliably detect OOD-examples in deep neural networks without introducing significant computational or complexity-wise overhead, we propose Distance to Modelled Embedding (DIME). The only assumptions underlying DIME are that we have access to the feature space of a trained deep neural network and samples from the training distribution. These simple assumptions allow us to derive a method that does not need any prior knowledge of the nature of OOD-examples nor is limited to classification classifications. Without needing to modify model training, DIME can be added after training with little extra work and computational overhead.We demonstrate that DIME consistently detect OOD-examples in multiple case studies, showing that it is a flexible tool for safer use of deep learning.

\section{Distance to Modelled Embedding}
We consider the problem of distinguishing between observations from distribution $P_\mathcal{X}$ and out-of-distribution (OOD) observations defined in data space $\mathcal{X}$. 
Depending of the data domain, the dimensionality of $\mathcal{X}$ may be large. 
We view a deep feed-forward neural network $\Phi: \mathcal{X} \rightarrow \mathcal{Y}$ as a sequence of non-linear transformations, where the transformation at layer $i$ is $\Phi_i: M_{i-1} \rightarrow M_i$ and $M_0 = \mathcal{X}$, mapping from $\mathcal{X}$ to the output domain. 
That is, for an observation tensor $ {x}_j \in \mathcal{X}$ the corresponding output $ {y}_j \in  \mathcal{Y}$ of a $m$-layer neural network is given by:
\begin{equation}
    \label{eq:network}
     {y}_j = \Phi({ {x}_j}) = \Phi_m\Big(\Phi_{m-1} \big(...\ \Phi_1( {x}_j) \big) \Big)
\end{equation}
Note that a transformation $\Phi_i$ may be as simple as a weight multiplication and bias addition in a multilayer perceptron, or a higher order operation such as a residual block in a residual neural network \cite{he2016deep}. The embedding vector of $ {x}_j$ in the feature space of an intermediate layer $i$ is then given by: 
\begin{equation}
    \label{eq:embedding}
    \Vec{\phi}_{j,i} = f_{\textrm{vector}} \bigg (\Phi_i\Big(\Phi_{i-1} \big(...\ \Phi_1( {x}_j) \big) \Big) \bigg)
\end{equation}
Where the transformation $f_{\textrm{vector}}: M_i \rightarrow \R^{p}$ is used to form a row vector representation of the tensor output of layer $i$. This transformation is necessary in, for instance, convolutional neural networks where the feature space has channel-dimensions as well as two spatial dimensions and is then typically implemented as the average across the spatial dimensions. Note that if $M_i = \R^{p}$, $f_{\textrm{vector}}$ may be the identity function.

We now introduce Distance to Modelled Embedding, DIME, that we use to detect OOD observations (see Figure \ref{fig:workflow} for a summary of the workflow). Given a training dataset $ {X}$ of length $n$ where each observation is assumed to be drawn from $P_{\mathcal{X}}$, we obtain an $n \times p $ embedding matrix $\Phi_{i, {X}} = [\Vec{\phi\ }_{1,i}^T\ ...\ \Vec{\phi\ }_{n,i}^T]^T$. We note that the rows of $\Phi_{i, {X}}$ are drawn from some probability distribution in $\R^p$, which according to the manifold hypothesis are expected to reside near a region of low dimensionality. But finding this region in a high-dimensional space may be challenging. We therefore use a simple and computationally efficient way to approximate said region based on the training set using an orthonormal set of basis vectors found by truncated singular value decomposition. We model the training set embedding as:
\begin{equation}
    \label{eq:svd}
    \Phi_{i, {X}} \approx \mU_k \mSigma_k \mV_k^T
\end{equation}
Where the columns of $n \times k$ matrix $\mU_k$ are the left singular vectors, $k \times k$ matrix $\mSigma_k$ contain the singular values, and the observations in $\Phi_{i,X}$ are projected onto the hyperplane spanned by the right singular vectors in the columns of $p \times k$ matrix $\mV_k$. 

This linear approximation provides a fixed description of the training set correlation structure in $\R^p$. We now view OOD-examples as violating this correlation structure. For an observation with vector embedding $\Vec{\phi}_{j,i}$, we measure this violation using DIME defined as the distance to the hyperplane in the embedding space as:
\begin{equation}
	\label{eq:residual_distance}
	\textrm{DIME}_{j,i} = \sqrt{(\Vec{\phi}_{j,i} - \hat{\Vec{\phi}}_{j,i})^2}
\end{equation}
That is, DIME is the reconstruction residual distance of the linear approximation $\Vec{\phi}_{j,i} \approx \hat{\Vec{\phi}}_{j,i} =\textrm{proj}_{\mV_k}\Vec{\phi}_{j,i}\mV_k^T$. 

To then determine if an observation is OOD in practice, we need to select a threshold distance. If we would assume that the rows in $\Phi_{i,X}$ are drawn from a multivariate normal distribution, it is well known that the squared residual distance follow a $\chi^2$-distribution. But this may be a too strong assumption, and we use the percentile compared to validation set distances instead to obtain a probability, $p_{ {x}_j \in P_{\mathcal{X}}}$, of observing an at most as long distance as the observed one as:
\begin{equation}
    p_{ {x}_j \in P_{\mathcal{X}}} = 1 - P_{\textrm{DIME}_{\textrm{Val}, i}}(\textrm{DIME}_{j,i})
\end{equation}
Where $P_{\textrm{DIME}_{\textrm{Val}, i}}: \R \rightarrow [0,1]$ is the observed percentile relative to all distances for observations in the validation set.

In summary, by using a straightforward approximation of the intermediate representations provided by neural networks, we can derive a computationally inexpensive one-step approach to detect out-of-distribution examples. Our method is permissive in terms of constraining model architecture or training, and is fully unsupervised in the sense that it does not require knowledge of what out-of-distribution examples may look like. 

\subsection{Selecting dimensionality}

The key hyperparameter of DIME is the dimensionality $k$ of the linear approximation in Eq. \ref{eq:svd}. We find it useful to select the dimensionality based on the ratio of explained training set variance. Given the full singular value decomposition of our $n \times p$ embedding matrix $\Phi_{i, {X}} = \mU \mSigma \mV^T$, with $n \times n$ left singular matrix $\mU$, $n \times p$ dimensional singular value matrix $\mSigma$ and $p \times p$ right singular matrix $\mV$ we calculate the vector of columnwise ratios of explained training set variances for each column in $\mV$ as:
\begin{equation}
    \label{eq:r2}
    \begin{split}
        R^2 &= \frac{\Vec{\sigma}^2}{\underset{i=1}{\overset{k_{max}}{\sum}} \Vec{\sigma}^2_i}\\
        \textrm{where}\ \Vec{\sigma}^2 &= \textrm{diag} \bigg(\frac{\mSigma \circ \mSigma}{n} \bigg)
    \end{split}
\end{equation}
Where $\circ$ denotes the element-wise product, $k_{max} = \min (n, p)$, $R^2 \in \R^{k_{max}}$, each element $R^2_i \in [0, 1]$ and $\sum_{i=1}^{k_{max}} R^2_i=1$. Assuming that the elements in $R^2$ are sorted from largest to smallest, we finally pick $k$ for cumulative explained variance ratio $r \in [0, 1]$ as:
\begin{equation}
    \label{eq:pick_k}
    k = \underset{\hat{k}} {\textrm{arg min}}\ \big\{ \hat{k} \in \{1, 2, ..., k_{max} \}\ | \ 
    r \leq \underset{i=1}{\overset{\hat{k}}{\sum}} R^2_i \big\}
\end{equation}
Selection based on the ratio of explained variance is arguably more intuitive than directly choosing dimensionality and allow for straightforward comparison between experiments where the rank of $\Phi_{i, {X}}$ may vary. It also allows us to explicitly control the degree of variation left unmodeled.

\subsection{Alternative formulations}

DIME is not the only way to measure distances within the feature space of a neural network. The Mahalanobis-distance is commonly used to detect outlier observations in data space. Given our network with training data embedding $\Phi_{i, {X}}$ and our embedded observation $\Vec{\phi}_{j,i}$, we can easily adapt the Mahalanobis distance for use in neural networks by calculating as:
 \begin{equation}
	\label{eq:simple_mahalanobis}
    d_{Mahalanobis, j, i} = \sqrt{
        (\Vec{\phi}_{j,i} - \Vec{\mu}_i)^T
        \mC^{-1}_{\Phi_{i, {X}}}
        (\Vec{\phi}_{j,i} - \Vec{\mu}_i)
    }
\end{equation}
Where $\mC_{\Phi_{i, {X}}}$ is the $p \times p$ empirical covariance matrix of $\Phi_{i, {X}}$ and $\Vec{\mu}_i$ is the $p$-dimensional vector of column-wise training set averages of $\Phi_{i, {X}}$. This formulation has been studied previously and showed promising results \cite{lee_unified_2018}. Compared to simpler metrics, such as the Euclidean distance, the Mahalanobis distance accounts for correlations between features in the target dataset and better accounts for that certain directions may have lower expected variance than others.

Alternatively, we can use our modelled embedding from Eq. \ref{eq:svd} and calculate the distance within the hyperplane. We again choose the Mahalanobis distance to account for different variances along different axes within the hyperplane, and formulate our distance within manifold as:
\begin{equation}
	\label{eq:mahalanobis}
    D_{within, j, i}=\sqrt{(\textrm{proj}_{\mV_k}\Vec{\phi}_{j,i})^T \mC_{\Phi_{i, {X}} \mV_k}^{-1} \textrm{proj}_{\mV_k}\Vec{\phi}_{j,i}}
\end{equation}
Where the projection $\textrm{proj}_{\mV_k}\Vec{\phi}_{j,i} = \Vec{\phi}_{j,i}\mV_k$, and $\mC_{\Phi_{i, {X}} \mV_k}$ is the $k \times k$ empirical covariance matrix of the $n \times k$ projection matrix $\Phi_{i, {X}} \mV_k$. A key difference to the simple Mahalanobis distance, is that  $\mC_{\Phi_{i, {X}} \mV_k}$ is guaranteed to be full rank avoiding a possibly singular covariance matrix. Another difference is that depending on the rank $k$ of our modelled, we can adjust the level of noise included in calculating the covariance matrix.

\section{Related Work}
In this section we introduce the methods we compare to DIME in the experiments below, namely softmax confidence by \cite{hendrycks2016baseline}, Monte-Carlo Dropout by Gal and Ghahramani \cite{gal_dropout_2016} and the Mahalanobis distance-based score described by Lee et al. \cite{lee_unified_2018}, which we denote as Deep-Mahalanobis. This section does not serve to give a complete overview of the field. The rationale for choosing softmax confidence is that due to its simplicity it is in our experience often encountered in deployed systems, Monte-Carlo Dropout is chosen since it is increasingly used in real-world applications \cite{janet2017predicting,gibson2018automatic,richardwebster2018psyphy,lotjens2019safe} and Deep-Mahalanobis due to its conceptual similarity to DIME and state-of-the-art performance. 

\subsection{Baseline - Softmax Confidence}

A simple baseline for OOD-detection is provided by Hendrycks and Gimpel \cite{hendrycks2016baseline} that observe that OOD-example predictions often have low confidence, where confidence is the maximum probability as predicted by a softmax-classifier.  Meaning that the confidence $c_j \in [0, 1]$ of observation $ {x}_j$ predicted by a $K$-way softmax-based classifier $\Phi_C: \mathcal{X} \rightarrow \R^K$ is defined as: 
\begin{equation}
    \label{eq:softmax_confidence}
    c_j = 
    \underset{i \in [1, K]}{\max} 
        \frac{e^{\Phi_C( {x}_j)_i}}
             {\underset{k=1}{\overset{K}{\sum}} e^{\Phi_C( {x}_j)_k}} 
\end{equation}
Where $\Phi_C( {x}_j)_k \in [0, 1]$ is the predicted probability that $ {x}_j$ belongs to class $k$. The confidence is then compared to a set threshold confidence to determine whether the observation is OOD or not. Assuming a well-calibrated model, the softmax confidence allows for a simple and interpretable metric for OOD-detection.

\subsection{Monte Carlo-Dropout}

Gal and Ghahramani proposed Monte Carlo-Dropout (MC-Dropout) \cite{gal_dropout_2016} , using prediction time dropout and Monte-Carlo sampling to produce a predicted probability distribution. 
The principles of MC-Dropout are based on the standard regularization method dropout \cite{hinton_improving_2012}. 
When training a neural network with a set of $L$ $n_{i+1} \times n_i$-dimensional weight matrices $\theta_i$ where each row $\theta_{i,r} \in \R^{n_i}$, for each iteration the elements of the weight matrices are dropped out according to:
\begin{equation}
    \label{eq:dropout}
        \hat{\theta}^T_i = \theta_i^T \cdot \textrm{diag}(\Vec{z}) \\
\end{equation}
Where each element $\Vec{z}_j \sim \textrm{Bernoulli}(p_i) \in \{0, 1\}$, $p_i \in [0, 1]$ is the dropout-probability of $\theta_i$ and $j \in [1, n_i]$.  Gal and Ghararami then illustrate how applying dropout during prediction time approximate Bayesian inference in deep Gaussian processes. If we now view the output of our neural network $\Phi(\cdot) = \hat{\Phi}(\cdot, \theta_1, ..., \theta_L)$ as a probability distribution, the model's expected prediction is simply given by the average over $M$ Monte-Carlo samples as:
\begin{equation}
    \label{eq:mc_dropout}
    \E[\Phi( {x}_j)] \approx \frac{1}{M} \underset{m=1}{\overset{M}{\sum}} \hat{\Phi}(x_j, \hat{\theta}_1, ..., \hat{\theta}_L)
\end{equation}
Where for each Monte-Carlo sample, a new set of weight matrices $\hat{\theta}_l$ are given by Equation \ref{eq:dropout}. For classification problems, the predictive uncertainty is then framed as the observation-wise entropy over Monte-Carlo samples. OOD-examples are then detected as high uncertainty observations.

\subsection{Deep-Mahalanobis}
DIME uses distances in the feature space of neural networks for OOD-detection and the closest prior art is the Mahalanobis-distance based score proposed by \cite{lee_unified_2018}, when used in a supervised setting reported state-of-the-art results on both OOD-detection and detection of adversarial attacks.  For each layer of a trained classifier the authors measure the Mahalanobis distance from training-set class centroids in the latent space of neural networks and then use the distance to the closest class centroid to separate in- and outliers. 

Given a trained softmax-based classifer with a set of $K$ classes, $n \times p$ training dataset embeddings $\Phi_{i, {X}}$ and an observation with  embedded feature vector $\Vec{\phi}_{j,i}  \in \R^p$ according to Eq. \ref{eq:embedding} the class with the closest centroid is found as:
\begin{equation}
	\label{eq:centroid}
    \Vec{\mu}_{c,i} = \underset{k}{\textrm{arg min}}\ ||\Vec{\phi}_{j,i} - \Vec{\mu}_{k,i}||_2
\end{equation}
Where the elements of class-centroid row vector $\Vec{\mu}_{k,i} \in \R^p$ are column-wise averages over every $\Vec{\phi}_{j,i}$ belonging to class $k \in K$. Then the class-centered Mahalanobis distance $D_{M}: \R^p \times K \times [1, 2, ..., m] \rightarrow \R$ is given by:
\begin{equation}
    \label{eq:class_centered_mahalanobis}
    D_{M, i}({x}_j, k, i) = \sqrt{
        (\Vec{\phi}_{j,i} - \Vec{\mu}_{k, i})^T
        \mC^{-1}_{\Phi_{i, {X}} - \mM_{k, i}}
        (\Vec{\phi}_{j,i} - \Vec{\mu}_{k, i})
    }
\end{equation}
Where $\mC_{\Phi_{i, {X}} - \mM_{k, i}}$ is the $p \times p$ empirical covariance matrix of $\Phi_{i, {X}} - \mM_{k, i}$ and $\mM_{k, i}$ is a $n \times p$ matrix where each row is the training set class embedding centroid $\Vec{\mu}_{k, i}$ for class $k$. 

To reach maximal performance, the observation $ {x}_j$ is then pre-processed into $\hat{ {x}}_j \in \mathcal{X}$ by adding a small perturbation of gradient noise according to:
\begin{equation}
    \label{eq:perturbation}
    \hat{ {x}}_j =  {x}_j - \epsilon \cdot \textrm{sgn} \bigg(
        \nabla_{ {x}_{j, i}} \big( D_{M, i}( {x}_j, c) \big)
    \bigg)
\end{equation}
Where $\nabla_{ {x}_{j, i}}: \R \rightarrow \mathcal{X}$  is the back-propagated gradient with respect to $ {x}_j$, $\epsilon \in \R$ is the magnitude of the perturbation, $\textrm{sgn}: \mathcal{X} \rightarrow \mathcal{X}$ the element-wise sign function, and $c \in K$ the closest class according to Equation \ref{eq:centroid}. Finally, a score of non-conformity that we denote Deep-Mahalanobis distance, $ d_{\textrm{Deep-Mahalanobis}, j, i} \in \R$, is calculated as the negative distance to the closest class-centroid to the embedded features the perturbed input $\hat{ {x}}_j$ as:
\begin{equation}
    \label{eq:deep_mahalanobis}
    d_{\textrm{Deep-Mahalanobis}, j, i} = - \underset{k \in K}{\min}\ D_{M, i}(\hat{ {x}}_{j, i}, k,i)
\end{equation}
In its original formulation, the Deep-Mahalanobis distance is calculated for each layer of the neural network by repeating Equations \ref{eq:centroid}-\ref{eq:deep_mahalanobis}. The confidence scores from each layer are then combined by weighted averaging with weights found using logistic regression between validation set and known OOD-example distances, meaning that it is supervised in contrast and assumes access to known OOD-examples for calibration. For the purpose of fair comparison, we restrict experimentation to the unsupervised Deep-Mahalanobis distance given by Equation \ref{eq:deep_mahalanobis}.

\section{Experiments}
In this section we describe our experiments that we use to evaluate how well DIME detect out-of-distribution examples. We compare our results to the baseline method of maximal softmax confidence, MC-Dropout and Deep-Mahalanobis. We limit our analysis to methods that does not strictly require any a priori knowledge of models of OOD-examples, and therefore use Deep-Mahalanobis without the original classifier. Similar to previous work \cite{ren2019likelihood}, we only use features from a single layer for OOD-detection for Deep-Mahalanobis instead of all layers. To allow proper ablation we also include simple Mahalanobis-distance directly in feature space according to Eq. \ref{eq:simple_mahalanobis}, class-centered Mahalanobis distance according to Eq. \ref{eq:class_centered_mahalanobis} as well as $D_{within}$ according to Eq. \ref{eq:mahalanobis}.

For DIME we vary the explained variance ratio of the linear approximation between \{0.9, 0.95, 0.99, 0.999, 1.0\} to select dimensionality according to Eq. \ref{eq:pick_k} and investigate its impact on performance (See Section \ref{section:dime_r2}), but only report performance for $r=0.99$ in comparison experiments. For Deep-Mahalanobis we vary the magnitude of the gradient noise, $\epsilon$ in Eq. \ref{eq:perturbation}, using the same values as the authors \cite{lee_unified_2018}, that is \{0, $10^{-2}$, $5 \cdot 10^{-2}$, $2 \cdot 10^{-3}$, $1.4 \cdot 10^{-3}$, $10^{-3}$, $5 \cdot 10^{-4}$\}, and report the best performance for each experiment. For MC-Dropout we train all models using dropout and report model uncertainty as the entropy of predictions of 30 Monte-Carlo samples using prediction time dropout.

Our experiments are inspired by Hendrycks and Gimpel \cite{hendrycks2016baseline} where OOD-example detection is framed as a binary classification problem. OOD-examples comprise different forms of random noise, manipulated test set data, data belonging to classes excluded from model training and data from other datasets. The success of detecting OOD examples is reported as precision-recall-curves (PR-AUC) separating test set examples from OOD examples. 

\subsection{Computer Vision}
\subsubsection{Fashion-MNIST}
\label{experiment:fashion-mnist}

To provide a simple classification problem with outliers encountered during prediction, we use the Fashion-MNIST \cite{xiao_fashion-mnist:_2017} dataset. Fashion-MNIST consists of 70 000 greyscale 28x28 pixel images, out of which 10 000 are test set images, of ten categories of fashion products. We excluded all three shoe classes (sandals, ankle boots and sneakers) from the training set. The intuition is that shoe-images should be detected as OOD-examples since all shoe-related information is absent from training data. All images, both in- and out-of distribution, were scaled to the range -1 to 1.

We trained a small CNN consisting of two ReLu-activated BN-Conv-BN-Conv-blocks with 32 and 64 3x3-filters respectively and 2x2 max-pooling with stride 1 after each followed by flattening and two 64-dimensional fully connected layers with 50 \% dropout. The model trained for 25 epochs using stochastic gradient descent with learning rate 0.01 and momentum 0.9 minimizing cross-entropy reaching a top-1 test-set accuracy of 90.2 \%. 

For OOD-examples, we follow Hendrycks and Gimpel \cite{hendrycks2016baseline} and use random noise of the same magnitude as pixel intensities (uniform and Rademacher), test set images either smoothed using a Gaussian kernel ($\sigma=2$ pixels) or randomly cropped, the excluded shoe images and test set images from MNIST \cite{lecun_mnist_nodate}. For DIME, Deep-Mahalanobis and the baseline method we did not use prediction-time dropout, but for MC-Dropout we used 30 Monte-Carlo samples per prediction. For DIME and Deep-Mahalanobis we used the output of the second convolutional block using the average over spatial dimension as $f_{vector}$ in Eq. \ref{eq:embedding} to provide features.

\subsubsection{CIFAR}
\label{experiment:cifar100}

To provide slightly more complex computer-vision problems, we use the CIFAR10 and CIFAR100-datasets \cite{krizhevsky2009learning} consisting of 60 000 32x32 pixel color images divided into 10- and 100-classes respectively. As OOD-examples, we used random uniform and Rademacher-noise, random mis-cropped (meaning that crops were padded with black resulting in broken images) and blurred test-set images, and images from the Describable Textures Dataset \cite{cimpoi2014describing} scaled into 32x32 pixel resolution.

On both datasets, we trained a 28x10-Wide ResNet \cite{zagoruyko2016wideresnet} for 200 epochs, batch size 128 and 30 \% dropout to minimize cross-entropy. As optimizer, we used stochastic gradient descent with an initial learning rate of 0.1 that was divided by 50 every 60 epochs, momentum 0.9 and $5\cdot10^{-4}$ weight decay. For data augmentation we used random horizontal flips. After training we achieved 81\% top-1 accuracy and 95.6 \% top-5 on CIFAR100 and 95 \% top-1 accuracy on CIFAR10. For OOD-detection, we evaluated all methods in the same manner as the Fashion-MNIST experiment using features from the last convolutional block for distance-based methods.

\subsection{Part-of-Speech Tagging}

To demonstrate the use of DIME to detect OOD-examples in another field than computer vision using another architecture than convolutional neural networks, we perform Part-of-Speech (POS) tagging on the the Penn Treebank corpus \cite{marcus1993building} using a bidirectional LSTM-based model \cite{hochreiter1997long}. The corpus was tokenized using default word tokenization in NLTK \cite{loper2002nltk}. A random subset of the corpus constituting 20 \% of all sentences was used as test set. The words were tokenized based on the training sentences (10063 words in vocabulary). We used all POS-tags present in the corpus (47 tags). All sequences of tokens were zero-padded to the length of the longest sequence in the training set (271 words).

The input tokens were embedded into a 128-dimensional vector which was fed into a single bi-directional LSTM-layer with 128 LSTM-cells. The token-wise LSTM-embeddings were then fed into a fully connected softmax classifier with 20 outputs. We used 20 \% dropout prior to the fully connected layer to enable MC-dropout even though we found it to reduce model performance. We trained the model for 25 epochs using the Adam optimizer \cite{kingma2014adam} with learning rate 0.001, batch size 64 while keeping 20 \% of the training sequences as validation data. Excluding padding tokens, the model achieved a token-wise accuracy of 89 \% on the test set.

For OOD-examples, we used tweets from the Ark-Tweet-NLP v0.3-dataset \cite{owoputi2013improved} and phrases of random words from the Penn Treebank-vocabulary as OOD-examples and performed token-wise OOD-detection, meaning that each token was treated as an individual observation. For distance-based methods we used the 128-dimensional token-wise embeddings from the LSTM-layer. Also for distance-based methods, we performed sequence-level OOD-detection, using the feature-wise maximum over the temporal dimension as $f_{vector}$.

Since the token-based model accept one-hot encoded input, we slightly modified Deep-Mahalanobis to modify the input based on the back-propagated gradient in word embedding space instead of token space. This modification was done since most modern deep learning frameworks provide implementations of token embedding-layers that typically only accept integer inputs. We did not see it necessary to reimplement embedding-layers to enable floating point modifications to token space-input to enable Deep-Mahalanobis that is ought to be used for any softmax-based classifier.

\section{Results}
\begin{figure}[t]
	\centering
    \includegraphics[width=\columnwidth]{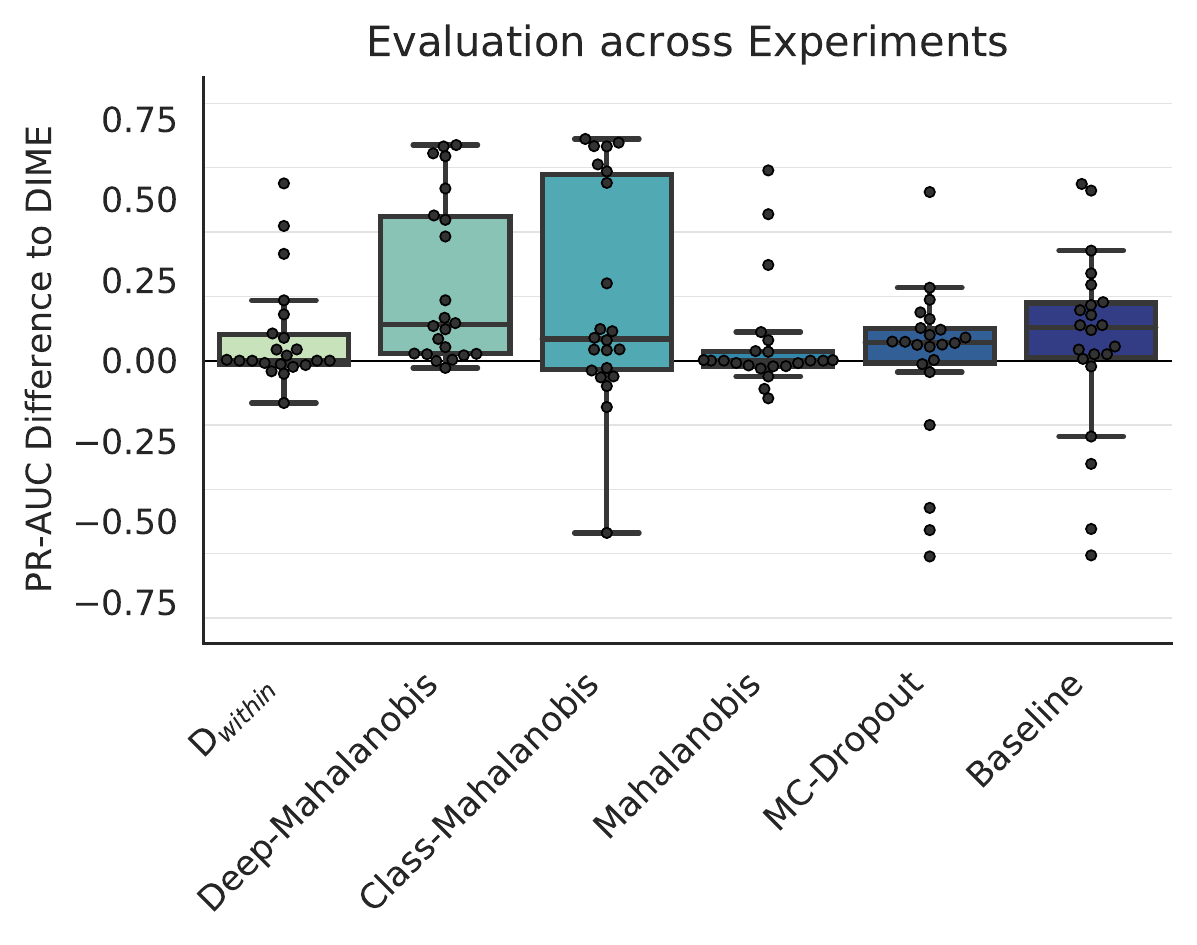}
    \caption{Box-plots showing PR-AUC relative to DIME across all experiments for each compared method. On the Y-axis is the difference in PR-AUC relative to DIME for each experiment, meaning that DIME performed better for each experiment above 0 and worse below. The median difference is shown as horizontal line on boxes, the upper and lower quartiles of differences as the boxes extent and the whiskers extend out to the closest point within an inter-quartile range from respective quartile.}
    \label{fig:across_experiments}
\end{figure}

Across all experiments and the methods investigated, DIME showed the best performance on average (see Figure \ref{fig:across_experiments} and Table \ref{table:across_experiments}). Since no method performed best in every single experiment and that the absolute scores varied depending on task difficulty, we chose to summarize our experiments relative to DIME across experiments and describe experiment results in more detail in sections below. This relative score is meaningful regardless of task difficulty compared to simply study the average performance across experiments. 

We only found statistically significant difference between DIME and Class- and Deep-Mahalanobis (see Table \ref{table:across_experiments}, using Wilcoxon signed-rank tests and significance at p $<$ 0.05) that both vary greatly in their relative performance. Both D$_{within}$ and simple Mahalanobis distance are both close to DIME in average performance, but DIME are up to 50 percentage units better in certain experiments while the opposite is not true. Both MC-Dropout and baseline softmax confidence are very close to DIME in average performance. But  MC-Dropout is requires use of dropout, which may not be optimal, and introduce large computational overhead due to sampling. On the other hand, both DIME and Simple Mahalanobis-distance are computationally efficient, but DIME is, as experiments below show, less sensitive to hyperparameters.

\label{section:results}

\begin{table}[t]
    \centering
    \caption{Average PR-AUC difference relative to DIME and P-value according Wilcoxon signed-rank test for signficant difference in rank means of paired samples.}

    \begin{tabular}{l  c  c }
        \hline
    	Method &  Average difference &   P-value \\
        \hline
      
        D$_{within}$      &               + 0.074         & 0.106 \\
        Deep-Mahalanobis  &               + 0.238         & $<$0.001 \\
        Class-Mahalanobis &               + 0.191         & 0.019 \\
        Mahalanobis       &               + 0.055         & 0.557 \\
        MC-Dropout        &               + 0.002         & 0.168 \\
        Baseline          &               + 0.060         & 0.088 \\
        \hline
    \end{tabular}
\label{table:across_experiments}
\end{table}

\begin{figure}[t]
    \centering
    \includegraphics[width=\textwidth]{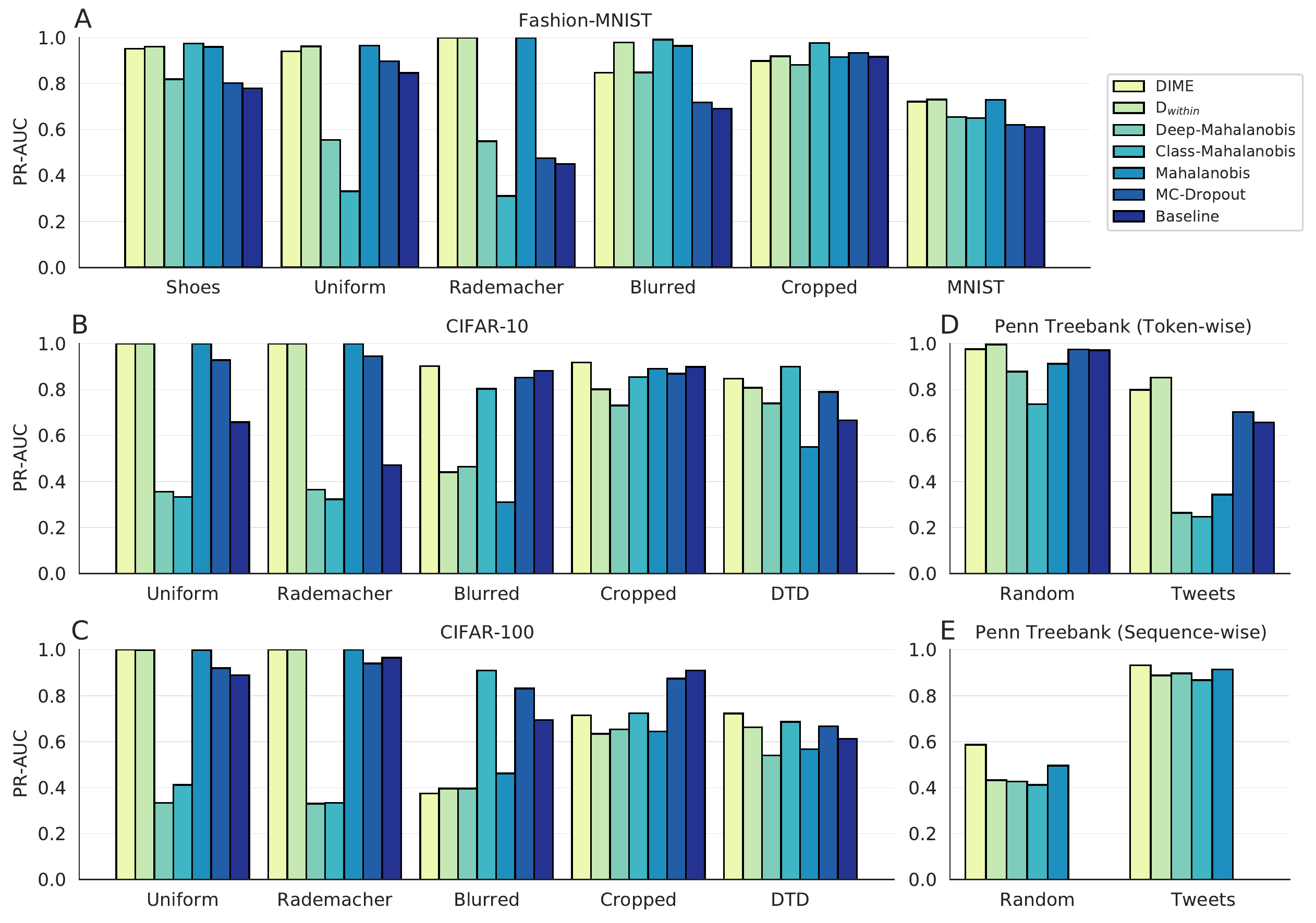}
    \caption{Detailed results of OOD-Detection in all experiments. On each Y-axis there is PR-AUC for OOD-detection, and on each X-axis the OOD-experiments used in each experiment. In the computer vision experiments, Fashion-MNIST (A), CIFAR-10 (B) and CIFAR-100 (C), results from the last convolutional block from each respective CNN are shown. For POS-tagging on the Penn Treebank-dataset, results from both token-wise (D) and sequence-wise (E) OOD-detection are shown.}
    \label{fig:results_all}
\end{figure}

\subsection{Computer Vision}

\subsubsection{OOD-detection on Fashion-MNIST is more challenging than expected}
For a small-scale dataset like Fashion-MNIST we might expect that most methods would perform equally well, but we observe a number of failure cases (see Figure \ref{fig:results_all}A). Class- and Deep-Mahalanobis completely fail to detect noise images on this dataset, as well as MC-Dropout and softmax confidence in the case of Rademacher-noise. Interestingly, most distance-based methods showed excellent detection for the excluded class (PR-AUC $>$ 0.95) whereas images from the MNIST-dataset were more challenging (0.61 $<$ PR-AUC $<$ 0.73). In this experiment, DIME perform consistently well for all datasets without any surprising failure cases.

\subsubsection{CIFAR-experiments confirm results from Fashion-MNIST experiment}
To large extent, we observe the same trends from the CIFAR-experiments as for Fashion-MNIST (see Figure \ref{fig:results_all}B and C). Again, Class- and Deep-Mahalanobis struggle to detect noise images. MC-Dropout and Baseline show most consistent results for blurred and cropped OOD-examples (PR-AUC $>$ 0.83 in all cases). No method performs best on every single dataset, but we conclude that DIME is consistently among the best performing methods with only exceptions being blurred and cropped images on CIFAR-100.

Comparing between CIFAR-10 and CIFAR-100, we observe a few differences. For instance, all methods find it more difficult to detect images from the DTD-dataset based and distance based methods perform worse on cropped on the CIFAR-100 experiment. In the case of blurred images, we also observe a drop in DIME performance from CIFAR-10 to CIFAR-100. Whether these difference arise from the different number of classes, or different degrees of converge of the trained models is something we leave for future work. 

\subsection{POS-tagging}

\subsubsection{Modelled embedding-based methods succeed on token-level OOD-detection}
DIME and D$_{within}$ both reliably detect both types of OOD-examples, and are to our knowledge the the first methods achieving PR-AUC $\ge$ 80 \% on the Tweets dataset in this task (see Figure \ref{fig:results_all}D). 
Surprisingly, distance-based metrics not using a modelled embedding perform worse on both datasets and fail completely on tweets. 
Both MC-Dropout and softmax confidence show fair results, and we observe better performance for softmax confidence on tweets compared to the original authors (66 \% compared to the original 41 \%) \cite{hendrycks2016baseline}. 
One difference that may explain this discrepancy is that they only trained their classifier on the Wall Street Journal-subset of Penn Treebank whereas we used the complete dataset. 
That model calibration is improved by training on more diverse datasets is consistent with literature \cite{hendrycks2019pretraining}.

Deep-Mahalanobis has previously only been evaluated on computer vision tasks and its performance does surprisingly not transfer directly to POS-tagging. Simple Mahalanobis-distance, class-centered Mahalanobis distance and Deep-Mahalanobis all perform worse compared to other methods on both random phrases and tweets. Out of those three, simple Mahalanobis-distance performs best on both OOD-datasets. The gradient push does in both cases improve upon the class-centered Mahalanobis indicating that our modification is not the source of the lower performance. These results suggest both that the modelled embedding is highly beneficial in this task and that class-centering may hurt performance despite POS-tagging being phrased as a token-wise classification problem. 

\subsubsection{Sequence-level detection using distance-based metrics is successful on tweets}

Distance-based metrics are straightforward to adapt for sequence-level OOD-detection despite that the model was trained for token-level POS-tagging. By simply aggregating the embedding features across the sequence we can use the distance-based methods in the same manner as for tokens. We pooled embedded features using max-pooling across the sequence and performed sequence OOD-detection on the same datasets used for POS-tagging (results in Figure \ref{fig:results_all}E).

All distance-metrics reliably detect tweets (PR-AUC $\ge$ 87 \%) and DIME showing best performance (PR-AUC = 93 \%), suggesting that distance-based sequence-level OOD-detection methods may be useful to detect distribution shift. Using this approach to detect phrases of random words was however much more challenging, and DIME was the only method with higher than 50 \% PR-AUC (59 \%). Using average pooling instead of max showed similar results on tweets, and slightly worse on random phrases (not shown). 

\subsection{Ablation Study}

\begin{figure}[t]
	\centering
    \includegraphics[width=\columnwidth]{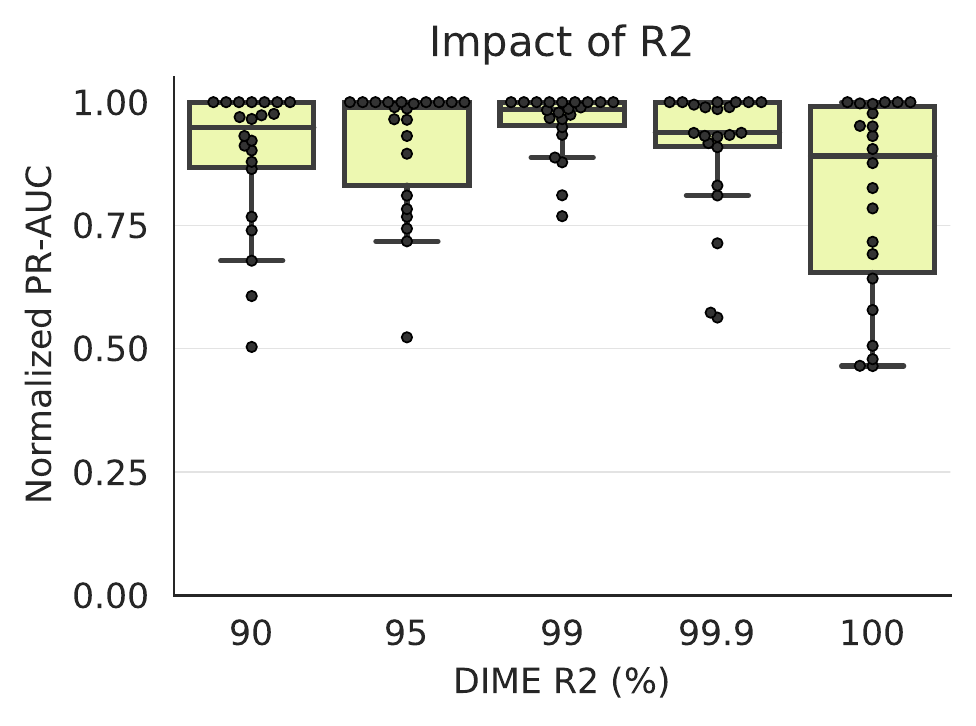}
    \caption{Box-plots showing DIME PR-AUC normalized across experiments. The results from each separate experiment is shifted so that the best performing configuration is equal to 1, and that the absolute difference to the other results within that experiment is kept unchanged. }
    \label{fig:dime_r2}
\end{figure}

\begin{figure*}[t]
	\centering
    \includegraphics[width=\textwidth]{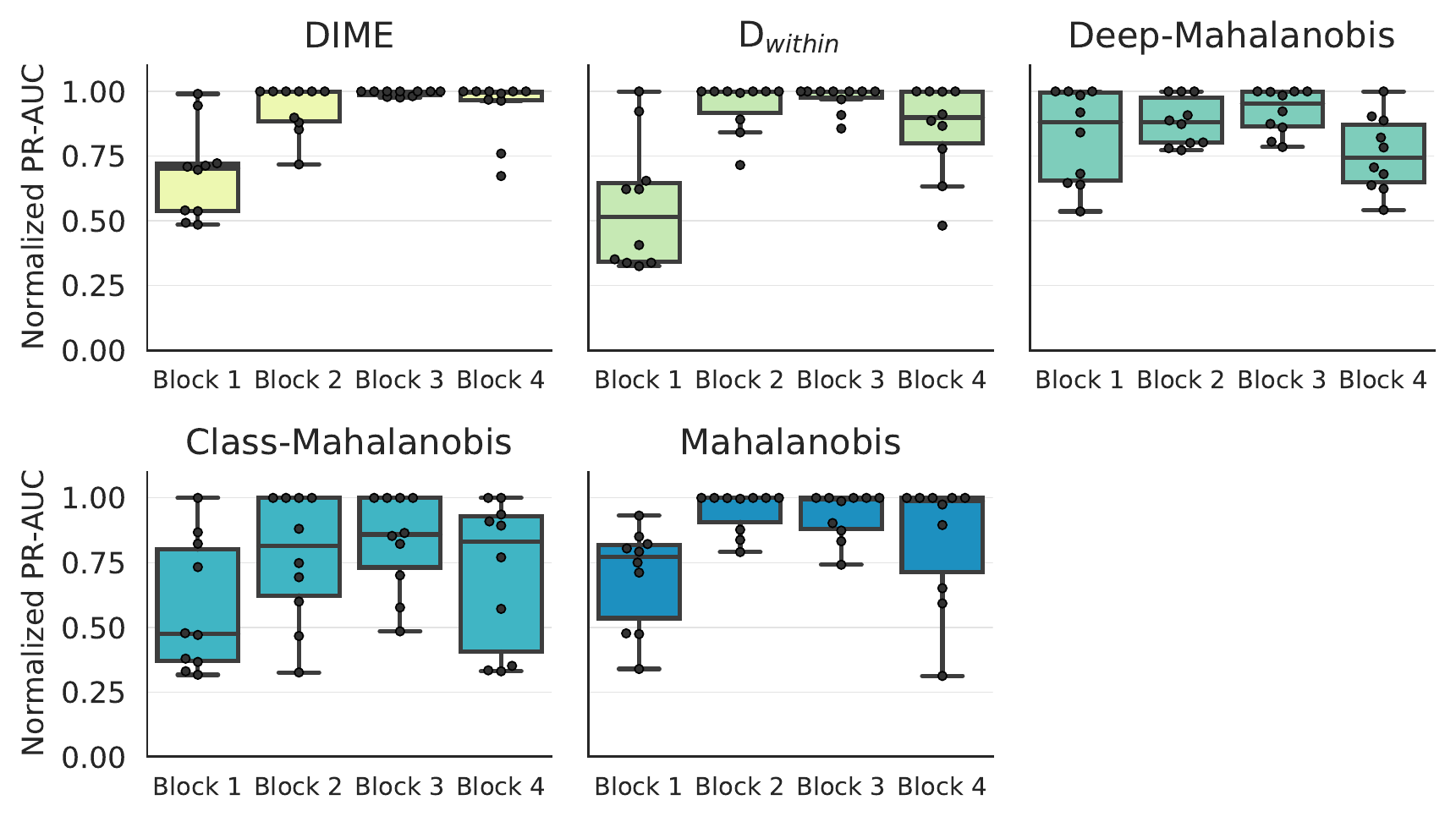}
    \caption{Box-plots showing PR-AUC normalized across feature depth of distance-based methods for OOD-detection in the CIFAR-10 and CIFAR-100 experiments. The results from each separate experiment is shifted so that the best performing configuration is equal to 1, and that the absolute difference to the other results within that experiment is kept unchanged. }
    \label{fig:cifar_blockeffect}
\end{figure*}

\subsubsection{Full-rank approximation is detrimental for DIME performance}
\label{section:dime_r2}
The rank of the linear approximation in Eq. \ref{eq:svd} is the key hyperparameter influencing the performance of DIME and we find it favorable to not use a full rank approximation (see Figure \ref{fig:dime_r2}). By normalizing PR-AUC values so that the best R2-value for each experiment results in a score of 1, and keeping the absolute difference to the other R2 intact, 99 \% explained variance show the highest median performance and lowest variance. Of particular notice is that using a full rank approximation performed worst on median, with very high variance in performance relative to truncated approximations. This may be due to overfitting in embedding space, and suggest that it is favorable to treat a small degree of variation in embedding space as noise for optimal performance in OOD-detection.

\subsubsection{DIME is least sensitive to feature depth}
Distance-based methods are influenced by which feature space is used to calculate distances. To investigate the degree of influence, we performed OOD-detection on both CIFAR-experiments using features from each wide-residual block in the trained models. For each method, we normalized PR-AUC values in the same manner as above (see Section \ref{section:dime_r2}) to visualize the sensitivity (See Figure \ref{fig:cifar_blockeffect}).  As expected does feature depth influence results, and in general are low-level features performing worst. An exception is that all methods perform best using features from the first block on random crops. Interestingly, all methods show best median performance using features from the second last rather than the last block. 

For the deep learning practitioner, low sensitivity to feature depth is important to avoid the need of rigorous evaluation for each use case. Based on our results, DIME is least sensitive to feature depth showing marginal difference between last and second last features. Importantly, DIME is less sensitive to feature depth than D$_{within}$ and simple Mahalanobis-distance that otherwise showed very similar average performance across experiments (see Figure \ref{fig:results_all} and Table \ref{table:across_experiments}).

\section{Discussion}
Deep learning is quickly being adopted in increasing numbers of applications and in order to use deep learning in safety-critical systems, it is important to know the limits of our models' knowledge. In this study, we propose  Distance to Modelled Embedding (DIME) as a new metric to detect out-of-distribution examples in neural network models. DIME is an unsupervised method for OOD-detection that impose minimal restrictions on the method or problem it may be applied. We simply assume that we have access to the training dataset and the feature space of our trained model as well as that the feature space can be approximated linearly sufficiently well. Given this, we then make a simple approximation of the region in feature space where our training data reside.  The simplicity of our formulation allows for straight-forward implementation in many use-cases, and introduces minimal additional complexity or computational overhead while performing at least on par with compared methods. 

This study is restricted to methods not strictly requiring access to OOD-examples for calibration. Assuming access to such observations hold true in certain applications, but in the general case this may not be true. In a well-studied field such as computer vision, we can relatively easy simulate certain types of OOD-observations but in other fields this may be challenging. Our work does however not exclude using OOD-exposure for calibration if OOD-examples are available, and we see the two approaches as complementary. Explicit use of OOD-examples is shown to greatly benefit OOD-detection \cite{hendrycks2018deep}, and for instance Deep-Mahalanobis has been combined with a modified loss-function to reduce confidence of OOD-examples \cite{papadopoulos2019outlier}. Since we observe that no method perform best throughout all experiments, we see it as beneficial to combine DIME with other methods for a more complete protection against OOD-examples. For classification problems, we may complement DIME using other low-overhead techniques such as re-calibration of confidence \cite{guo2017calibration}. For other objectives, such as regression, we may require adopted training regimes using for instance a Bayesian approach to attain predictive uncertainty. Based on our results, MC-Dropout performs well for OOD-detection, despite criticism of its correctness \cite{osband_deep_2016}. But compared to DIME, a major downside of MC-dropout, and other Bayesian approaches, is that they require modified training procedures, in this case use of dropout. This limits their use, and may introduce a trade-off between model performance and possible uncertainty estimation OOD-detection whereas DIME is applicable in any model. Another downside of the Bayesian approaches is that they introduce large computational overhead due to sampling. DIME, on the other hand, introduce negligible overhead making it applicable even in environments with limited computational resources.

We were surprised by the poor performance of Deep-Mahalanobis in our experiments. One reason could be that in contrast to its original use, we modified it to not rely on an explicit classifier calibrated against known OOD-examples, as well as only using feature from a single layer instead of all layers as originally formulated. The motivation behind our modifications are that we aim to study all distance-based methods under similar circumstances that we see more likely to be implemented in practice. In our formulation, simple Mahalanobis distance in feature space consistently perform better than both class-centered Mahalanobis distance and Deep-Mahalanobis. The class-centering also impose a critical limitation of the approach by limiting its use to classification problems. In certain experiments, we saw better performance using this approach but this was not consistently true and surprisingly not true in POS-tagging despite it being framed as a classification problem. The second major limitation of this approach is that the gradient push used to increase performance introduce new hyperparameters that are non-trivial to tune. In our experiments, we chose to vary the magnitude of the gradient and only report the best performing one. In practice we would however need to choose a magnitude to use. Something that may be difficult without access to OOD-examples. Similarly, it also require properly tuned gradient initialization that needs to be adjusted on a case-by-case basis. Thirdly, by using back-propagation, Deep-Mahalanobis introduce significant computational overhead (a limitation shared with other methods \cite{liang2018enhancing}. This overhead is a significant hurdle when implementing OOD-detection in systems with time- or hardware-constraints. Distance calculation without back-propagation, for instance using DIME, introduce negligible computational overhead compared to running model inference, even when OOD-detection is run on CPU.

We explored further alternative formulations of DIME that showed negative results in preliminary experiments and we decided to not include them in this work. For instance, we attempted several methods to merge the residual distance with D$_{within}$ for instance using the L2-norm, with and without normalization, of the two. As well as maximal, minimal or multiplied probabilities acquired from percentile binning of the two distances. We also explored using autoencoders, both regular and variational, to model the embedded features but did not see beneficial results compared to the linear approximation in Eq. \ref{eq:svd}. For future research, we instead suggest exploring the use of DIME beyond that of stationary models. Bayesian methods are for instance used in reinforcement learning to promote exploration \cite{osband_deep_2016,gal_dropout_2016} or enable safer decision making \cite{lotjens2019safe,da2020uncertainty}. We hypothesize that methods for OOD-detection are complementary to predictive uncertainty also in reinforcement learning, and believe that they can help improve active systems.

\section{Conclusion}
Reliable detection of out-of-distribution examples is an important component for deployment of deep neural networks in safety-critical systems. We present our metric Distance to Modelled Embedding (DIME) that we use to detect out-of-distribution examples during prediction time. By linearly approximating the training dataset in feature space, we can derive a simple, unsupervised, highly performant and computationally efficient method. DIME allows us to add prediction-time detection of out-of-distribution examples to neural network models without altering architecture or training. Our case studies show that DIME used as an add-on after training reliably matches state-of-the-art performance while being highly versatile and only introduce negligible overhead, demonstrating its benefits for safer use of deep learning. 

\section*{Acknowledgements}

The authors would like to thank Dr. Olivier Cloarec at Sartorius Corporate Research for fruitful discussions on matters regarding multivariate statistics and for providing valuable feedback. The authors would also like to thank Dr. Kimin Lee at UC Berkeley for providing guidance on implementation details of Deep-Mahalanobis.

\bibliographystyle{unsrt}  

\bibliography{bibliography.bib}   

\end{document}